\documentclass[runningheads]{llncs}
\usepackage[T1]{fontenc}
\usepackage{graphicx}
\usepackage{color}
\usepackage{hyperref}

\usepackage{booktabs, tabularx}
\usepackage{multirow}
\usepackage{threeparttable}
\usepackage{threeparttablex}

\usepackage{amsmath,amsfonts,amssymb}

\usepackage{algorithmic}
\usepackage{algorithm}
\usepackage{array}
\usepackage{graphicx}
\usepackage{rotating}
\usepackage[table]{xcolor}

\usepackage{adjustbox}
\usepackage{multicol}
\usepackage{caption}
    \newcommand{\midsepremove}{\aboverulesep = 0mm \belowrulesep = 0mm}
    \midsepremove
    \newcommand{\midsepdefault}{\aboverulesep = 0.1mm \belowrulesep = 0.1mm}
    \midsepdefault
 \newcommand{\notesize}{\fontsize{7}{10.5}\selectfont}

\begin{document}
\title{Leveraging Duration Pseudo-Embeddings in Multilevel LSTM and GCN Hypermodels for Outcome-Oriented PPM\thanks{This work has been partially supported by the European Union under the NextGenerationEU initiative through the REPA (aRtificial Intelligence for Process Analytics) project, Grant Assignment Decree No. 2022CJWPNA by the Italian Ministry of Ministry of University and Research (MUR).}}
\titlerunning{Duration Pseudo-Embeddings in Multilevel Hypermodels}
%
\author{Fang Wang\inst{1}\orcidID{0000-0002-7349-8888} \and
Paolo Ceravolo\inst{2}\orcidID{0000-0002-4519-0173}\and
Ernesto Damiani\inst{1}\orcidID{0000-0002-9557-6496}}
\authorrunning{F. Wang et al.}
%
\institute{College of Computing and Mathematical Sciences, Center for Cyber-Physical Systems, Khalifa University, Abu Dhabi, UAE \and Computer Science Department, University of Milan, Milan, Italy}
\maketitle              
\begin{abstract}
Existing deep learning models for Predictive Process Monitoring (PPM) struggle with temporal irregularities, particularly stochastic event durations and overlapping timestamps, limiting their adaptability across heterogeneous datasets. We propose a dual input neural network strategy that separates event and sequence attributes, using a duration-aware pseudo-embedding matrix to transform temporal importance into compact, learnable representations. This design is implemented across two baseline families: B-LSTM and B-GCN, and their duration-aware variants D-LSTM and D-GCN. All models incorporate self-tuned hypermodels for adaptive architecture selection. Experiments on balanced and imbalanced outcome prediction tasks show that duration pseudo-embedding inputs consistently improve generalization, reduce model complexity, and enhance interpretability. Our results demonstrate the benefits of explicit temporal encoding and provide a flexible design for robust, real-world PPM applications. \footnote{The implementation code for LSTM hypermodels and pseudo-embedding matrix is available at \url{https://github.com/skyocean/HyperLSTM-PPM}}

\end{abstract}

\label{Intro}
\section{Introduction}
Outcome-oriented Predictive Process Monitoring (PPM)~\cite{ceravolo2024predictive} is a key AI application in Business Process Management, aiming to forecast outcomes of ongoing cases using historical event logs~\cite{ceravolo2024tuning}. Deep learning models, especially recurrent neural networks (RNNs) and Graph neural networks (GNNs)~\cite{camargo2019learning,wang2025hgcnoselftuninggcnhypermodel}, have significantly improved PPM by capturing temporal and structural dependencies. Yet, existing methods face three core limitations.

First, they struggle with temporal irregularities, such as stochastic event durations and overlapping timestamps~\cite{ceravolo2024tuning,fracca2022estimating}. Second, fixed architectures and static hyperparameters hinder generalization across diverse datasets. Third, outcome-oriented PPM remains underexplored~\cite{ceravolo2024predictive,bellandi2021correlation,weinzierl2021exploring} with many approaches still relying on traditional machine learning models~\cite{aljebreen2024predicting,teinemaa2019outcome}, lacking expressiveness and scalability. A key challenge lies in modelling temporal heterogeneity effectively without increasing complexity or overfitting. Many current models rely on hand-crafted features or flatten event traces, blurring event-level and case-level semantics.

To address these issues, we propose a dual-input framework that explicitly models event- and case-level attributes. At its core is a duration-aware pseudo-embedding matrix that encodes the relative importance of event durations into compact, learnable representations. This framework is instantiated in two baseline families, B-LSTM (recurrent) and B-GCN (graph-based), and extended to duration-aware variants, D-LSTM and D-GCN, by incorporating temporal pseudo-embeddings. Our contributions are threefold. First, we propose a duration-aware pseudo-embedding strategy that captures temporal irregularities directly from timestamps, thus eliminating the need for manual feature engineering. Secondly, we design self-tuned hypermodels for both LSTM and GCN variants, which enable automatic adaptation of the architecture to the characteristics of the dataset and the granularity of the input. Finally, we empirically demonstrate that our duration-aware models match or exceed baseline performance on balanced and imbalanced benchmark datasets while offering better interpretability and maintaining leaner architectures. This modular, temporal informed design advances adaptive PPM by providing a framework that considers the temporal aspects of the problem and incorporates modular components that can be adapted as necessary.

The rest of the paper is structured as follows: Section 2 introduces the architecture and pseudo-embedding strategy. Section 3 covers datasets and experiments. Section 4 presents results, and Section 5 concludes.
\vspace{-0.3cm}
\section{Model Architectures and Representations}
\label{sec:model_architecture}
In the PPM literature, a \textit{case} is a sequence of events associated with a specific process instance, as recorded in an event log. When multiple cases exhibit identical sequences of events, these recurring patterns are referred to as \textit{traces}.

This section introduces two families of predictive models that are commonly used for event sequence modelling: \textit{Graph Convolutional Networks (GCNs)} and \textit{Long Short-Term Memory Networks (LSTMs)}. For each family, we consider two variants: a baseline model and a duration-aware extension.

Both GCN and LSTM models adopt a multi-level input architecture that encodes information at two distinct levels: (i) \textbf{Event-level (node-level):} capturing attributes of individual events such as activity type, timestamp, and resource; (ii) \textbf{Case-level (graph-level):} capturing contextual or aggregated attributes of the entire case. We first outline each model’s structure and data representation strategy, then introduce a shared duration-aware component: the \textit{duration-aware pseudo-embedding}. This mechanism enriches event-level representations with temporal features, allowing the models to better capture process dynamics.

\vspace{-0.3cm}
\subsection{Graph Convolutional Network Models}
\label{sec:gcn_model}
\subsubsection{Graph-Based Representation}
Each case is represented as a directed, weighted graph $\mathcal{G}_{G_j}$, where nodes $X_i$ represent individual events and edges capture temporal transitions. Node attributes include an activity label $\mathcal{A}_i$, universal attributes $U_i$, and conditional (specific) attributes $B_i$, collectively encoded into a composite vector $\mathbf{v}_{N_i} = [\mathcal{A}_i, B_i, U_i]$. Duration is computed from start and end timestamps of each event and treated as a universal attribute. Attributes are encoded using one-hot (categorical) or min-max scaling (numerical). Missing or irrelevant specific attributes are masked using median imputation (numerical) or a padding token (categorical, encoded as $-1$). At the graph level, each case $G_j$ is associated with global attributes $F_j$, encoded as a vector $\mathbf{v}_{G_j}$. The resulting graph is defined by: \textit{node matrix:} $\mathbb{V}_{N(G_j)} \in \mathbb{R}^{n \times d_N}$; \textit{edge index tensor:} $\mathbb{E}_{G_j} \in \mathbb{R}^{2 \times (n-1)}$, connecting each $X_i \rightarrow X_{i+1}$; \textit{edge weight vector:} $\mathbb{W}_{G_j} \in \mathbb{R}^{n-1}$. Edge weights $w_{(i \rightarrow i+1)} = T^s_{i+1} - T^s_i$ represent time gaps between consecutive event start times, normalized via min-max scaling. This captures temporal dependencies while accounting for simultaneous events (where $w=0$). Unlike traditional end-time-ordered approaches, our method sequences events by start time to maintain temporal causality. This explicitly preserves cases where early-starting events with extended durations overlap subsequent events, with duration attributes encoded in node vectors to resolve scheduling ambiguities.

\subsubsection{Baseline GCN Architecture(B-GCN)}  
The baseline GCN (B-GCN) model processes graph structured event logs using edge weight aware graph convolutions. Each graph $\mathcal{G}_{G_j}$, containing node attributes $\mathbb{V}_{N(G_j)}$ and temporal edge weights $\mathbb{W}_{G_j}$, is fed through a stack of \textbf{GCNConv} layers. These layers aggregate information from local neighborhoods using weighted message passing, allowing the model to capture both structural and temporal dependencies.

We adopt the GCNConv variant that supports edge weights, extending the classical formulation by \cite{kipf2016semi} to allow for time-sensitive propagation. The graph convolution layer updates node attributes using the normalized weighted adjacency matrix:
\begin{equation}
\mathbb{V}^{(l+1)} = \sigma\left( \tilde{D}^{-\frac{1}{2}} \tilde{A}_w \tilde{D}^{-\frac{1}{2}} \mathbb{V}^{(l)} W^{(l)} \right),
\end{equation}
where $\tilde{A}_w = A_w + I$ includes edge weights and self-loops, and $\tilde{D}$ is its degree matrix. $\sigma$ is a non-linear activation and $W^{(l)}$ is a learnable weight matrix. This normalization stabilizes learning across varying graph sizes and node degrees. After $L$ convolutional layers, node embeddings $\mathbb{V}^{L}_{N(G_j)}$ are aggregated via global pooling into a graph-level vector $\mathbf{z}_{G_j}$. In parallel, the case-level attributes $\mathbf{v}_{G_j}$ are processed through dense layers to yield $\mathbf{v}^d_{G_j}$. The final trace representation is formed as: $\mathcal{Z}_{G_j} = \textit{Concatenate}(\mathbf{z}_{G_j}, \mathbf{v}_{G_j}^d)$. This representation is then passed through fully connected layers to produce the prediction $\hat{y}_{G_j}$.

\subsection{LSTM Network Models}
\label{sec:lstm_model}
\vspace{-0.2cm}
\subsubsection{Sequential Representation Construction}
For the LSTM models, each eventlog is modeled as an ordered sequence of event vectors. Each event $X_i$ within a trace $G_j$ is represented by a preprocessed feature vector $\mathbf{v}_{N_i} \in \mathbb{R}^{d_N}$, consistent with the node vector used in the GCN variant. These vectors are stacked in order of their event timestamps to form a matrix $\mathbb{S}_{G_j} \in \mathbb{R}^{n \times d_N}$, where $n$ is the number of events in the trace. The ordered event sequence $\mathbb{S}_{G_j}$ serves as LSTM input, preserving temporal order. To support overlapping events, we augment each event vector with the time difference flag between consecutive start timestamps. Case-level attributes $\mathbf{v}_{G_j}$ are encoded separately and incorporated later during the output stage of the model. Padding and masking are applied to ensure uniform input shapes across variable-length sequences.

\subsubsection{Baseline LSTM Architecture (B-LSTM)}
The baseline LSTM(B-LSTM) model consumes the event sequence matrix $\mathbb{S}_{G_j}$ and processes it through a multi-layer LSTM network to capture temporal dynamics. The output of the final LSTM hidden state is taken as the trace-level representation $\mathbf{z}_{G_j}$. In parallel, the separately encoded case-level attributes $\mathbf{v}_{G_j}$ are passed through fully connected layers to obtain $\mathbf{v}_{G_j}^d$. Similar to GCNs, these two components are concatenated to form the combined representation:~$\mathcal{Z}_{G_j}$, which is passed through dense prediction layers to produce the final output $\hat{y}_{G_j}$. 

\begin{algorithm}[htb!]
\caption{Pseudo-Embedding Duration Bin Matrix}
\label{alg:pseudo_embedding}
\begin{algorithmic}[1]
\REQUIRE Set of activities (nodes) $\{\mathcal{A}_i\}$; duration values $\{T^d_i\}$ for each $\mathcal{A}_i$; set of graphs $\{G_j\}$.
\ENSURE Pseudo-embedding matrices where each $\mathcal{A}_i$ is represented as vector $\mathbf{v}_{bin_i}$ for all $G_j$.

\STATE Initialize cut-off value $T_{\text{cut}}$ and number of quantile bins $N_{\text{quant}}$.

\REPEAT
    \FOR{each event $\mathcal{A}_i$}
        \IF{$T^d_i < T_{\text{cut}}$}
            \STATE Assign $T^d_i$ to a unique bin $b$.
        \ELSE
            \STATE Calculate quantile bins $\{\bar{b}\}$ based on $N_{\text{quant}}$.
            \STATE Remove duplicates and adjust $\{\bar{b}\}$ to cover the full duration range.
            \STATE Assign $T^d_i$ to the appropriate quantile bin $\bar{b}$.
        \ENDIF
        \STATE Assign final duration bin $b'_i$ to $\mathcal{A}_i$.
    \ENDFOR

    \STATE Calculate bin frequencies $\{f_b'\} = \{f_b, f_{\bar{b}}\}$.

    \IF{bin frequencies $\{f_{\bar{b}}\}$ are balanced (within threshold)}
        \STATE \textbf{break}
    \ELSE
        \STATE Update $T_{\text{cut}}$ and $N_{\text{quant}}$ accordingly.
    \ENDIF
\UNTIL{stopping condition is met}

\STATE Extract unique pairs $(\mathcal{A}_i, b'_i)$.
\STATE Treat each pair $(\mathcal{A}_i, b'_i)$ as a term in the corpus.

\FOR{each graph $G_j$}
    \STATE Treat $G_j$ as a document.
    \STATE Construct a TF-IDF matrix with rows as $\mathcal{A}_i$ and columns as bins $b'_i$:
    \FOR{each term $(\mathcal{A}_i, b'_i)$}
        \STATE Compute $\text{tf-idf}(\mathcal{A}_i, b'_i)$.
    \ENDFOR
\ENDFOR

\RETURN Pseudo-embedding matrices $\mathbf{v}_{bin_i}$ for each $\mathcal{A}_i$ across all $G_j$.
\end{algorithmic}
\end{algorithm}
\setlength{\textfloatsep}{0pt}

\subsection{Duration-Aware Pseudo-Embedding Augmentation}
\label{sec:pseudo_embedding}
To enrich event level representations with duration-aware temporal context, we augment raw node/event inputs with a pseudo-embedding matrix derived from discretized duration attributes. These embeddings provide an auxiliary feature space that complements standard activity and attribute encodings. Following \cite{wang2025comprehensive}, we employ a hybrid binning strategy: durations below a fixed cut-off receive unique bins, while longer durations are grouped using quantile-based binning. This balances bin frequency while preserving granularity for short events.

Each event is then mapped to a (activity, duration-bin) pair, treated as a term in a TF-IDF-inspired scheme. Cases are treated as documents in a corpus, and we compute term relevance by weighing duration-bin frequency within a case against its global rarity. This yields a pseudo-embedding matrix encoding how distinctively each activity-duration combination characterizes a case. The full procedure is summarized in Algorithm~\ref{alg:pseudo_embedding}. These embeddings are processed in parallel with raw inputs through separate GCN or LSTM layers and later fused to form a richer case representation. 

This mechanism is integrated into both model families via a parallel encoding pathway, forming two new models- Pseudo-Embedding Duration LSTM (D-LSTM) and Pseudo-Embedding Duration GCN (D-GCN).  Each case/graph has two input branches: one for the standard event/node attributes and another for the duration-based pseudo-embedding vectors. Both branches are processed independently through identical encoder layers (either GCN or LSTM). The resulting hidden representations are then concatenated and passed through additional GCN or LSTM layers, followed by pooling (in the GCN case), fusion with case level attributes, and final prediction layers. This design allows each input stream to be encoded in its own latent space before joint modeling, enabling the architecture to leverage both raw and duration-augmented perspectives.

\subsection{HyperModel Configuration}
We develop four hypermodels spanning GCN and LSTM architectures, each automatically tuned per dataset.\textbf{Common hyperparameters} across all models include: learning rate and schedule; optimizer type and configuration; batch size and loss function; Dropout and batch normalization options; dense layer size and activation function. \textbf{GCN-specific hyperparameters} include number of GCN layers and size and pooling operations. \textbf{LSTM-specific hyperparameters} include number of LSTM layers and hidden units and regularization strength. We handle masked feature values (encoded as $-1$) during preprocessing to prevent bias. Table \ref{tab:hyperparameters} details all parameter types and ranges. \footnote{Range and sub-parameters for each specific optimizer and learning rate scheduler are also tuned.}
\vspace{-0.95cm}
\begin{table}[htb!]
\centering
\caption{Hyperparameters and Their Tuning Ranges/Types}
\label{tab:hyperparameters}
\begin{tabularx}{\linewidth}{l@{\hspace{1pt}}X}
\hline
\small
\textbf{Hyperparameter}          & \textbf{Range}                            \\\toprule
\multicolumn{2}{p{\hsize}}{\textit{\textbf{Node/Event Level Layers}}}
\\
Number of layers     & GCN: 1-5;    LSTM: 1-3\\                                
Hidden Units                    & 16-512 \\
Dropout      & flag: Y/N; rates: 0.2-0.7\\
Batch norm & flag: Y/N; momentum: 0.1-0.999; eps:1e-5-1e-2 \\
LSTM L2 & 1e-5-1e-2\\ 
GCN Activation & ReLU, Leaky\_ReLU, ELU, Tanh, Softplus, GELU\\
GCN Skip Connection & Y/N\\\midrule
\textit{\textbf{GCN Pooling}} & mean, add, max\\\midrule
\multicolumn{2}{p{\hsize}}{\textit{\textbf{Dense Layers}}}\\
Number of layers     & 1-3   \\                                
Dense Units                    & GCN: 16-512; LSTM: 16-256 \\
Dropout      & flag: Y/N; rates: 0.2-0.7\\
Batch norm & flag: Y/N; momentum: 0.1-0.999; eps:1e-5-1e-2 \\ 
Activation & ReLU, Leaky\_ReLU, ELU, Tanh, Softplus, GELU(GCN)\\
LSTM L2& 1e-5-1e-2\\\midrule
\multicolumn{2}{p{\hsize}} {\textit{\textbf{Optimizer and Learning Rate Scheduler}}}\\
Learning Rate & 1e-5-1e-2 (log)\\
Weight Decay & 0-1e-3\\
L1 & 0-1e-3\\
Optimizers & Adam, RMSprop, SGD\\
Schedulers & GCN: Step, Exponential, Reduce-on-Plateau, Polynomial, Cosine Annealing, Cyclic, One Cycle\\
&LSTM: Exponential, Inverse Time, Piecewise Constant, Polynomial\\ \midrule
\textit{\textbf{GCN Loss Function}} & CrossEntropy, MultiMargin\\\midrule
\textit{\textbf{Batch Size}} & 16, 32, 64, 128, 512 \\\bottomrule
\end{tabularx}
\end{table}
\vspace{-0.8cm}

\section{Experiments}
\subsection{Datasets and Processing}
To evaluate model robustness and generalization under different data conditions, we use two representative datasets: one \textbf{highly imbalanced} and one \textbf{balanced}. Specifically, we employ the synthetic \textbf{Patients} dataset to simulate extreme class imbalance, and the real-world \textbf{BPIC12} benchmark \cite{BPIC12set} to represent balanced classification tasks.

Patients (Synthetic Healthcare Data) dataset contains 2,140 cases representing patient interactions within a healthcare system. It includes 3 numerical and 1 categorical attribute at the case level, and 3 numerical plus up to 3 categorical attributes (including one universal) at the event level. Each case is labeled with one of five possible outcomes. The label distribution is highly skewed: the majority class accounts for 40.74\% of cases, and the minority just 1.12\%, yielding an imbalance ratio of approximately 36:1. This setup makes it ideal for testing model performance under class imbalance and attribute heterogeneity.

BPIC12 (Loan Application Event Log) captures real-world loan and overdraft application processes in a multinational financial institution. Each case ends with one of three outcomes: \emph{accepted}, \emph{declined}, or \emph{canceled}. We use a curated version with balanced class sizes (2,224 cases per class). It includes one numerical attribute at the case level and two universal categorical attributes at the event level. BPIC12 introduces temporal modeling challenges due to frequent \emph{timestamp collisions}—i.e., multiple events with identical start times \footnote{The start time and complete time of an activity is based on its transition status.}—which are well-suited for evaluating time-aware representations.

As summarized in Table~\ref{tab:data}, the datasets differ substantially in case length, attribute complexity, and temporal structure. \textit{Patients} features shorter but more heterogeneous traces, while \textit{BPIC12} has longer cases with simpler event attributes but denser temporal overlap. Case lengths range from 12 to 77 events, introducing varying degrees of temporal dynamics and representational demands.

All events are timestamped with start and completion times. Event durations are calculated as the difference between start and end times, rounded to the nearest minute. In \textit{Patients}, durations under 5 minutes are assigned unique bins, while longer durations are grouped into 24 quantile-based bins. In \textit{BPIC12}, durations are discretized into two bins: zero and non-zero duration. These bin labels feed into the pseudo-embedding module used for temporal augmentation. 
\vspace{-1cm}
\begin{table}[htb!]
\centering
\setlength{\tabcolsep}{3pt}
\caption{Statistics of the datasets used in the experiments}
\label{tab:data}
\notesize
\begin{tabularx}{\linewidth}{cccccccccccX}
\hline
data & \#Case  & max-len  & min-len  & med-len & \multicolumn{3}{c}{\#Event Attr \& Size } & \multicolumn{3}{c}{\#Case Attr \& Size} & \# O\\
\cmidrule[1pt](lr){3-5}\cmidrule[1pt](lr){6-8}\cmidrule[1pt](lr){9-11} 
\midrule
    BPI12 & 6672 & 77   & 12   & 18   & (N)2    & (C)0    & -    & (N)1    & (C)0    & -    & 3 \\
    Patients & 2140 & 9    & 4    & 7    & (N)3    &(C)3    & [10,3,3] & (N)3    & (C)1    & 2    & 5 \\
\bottomrule
\end{tabularx}
\end{table}
\setlength{\textfloatsep}{0pt}
\vspace{-0.8cm}
\subsection{Hyperparameter Search}
We employ two search strategies tailored to our model families and dataset characteristics. For LSTM-based HyperModels, we use the Hyperband algorithm, selected for its efficiency in large search spaces through early-stopping-based exploration. Search objectives differ by dataset type: for balanced datasets, validation accuracy is maximized; for imbalanced datasets, we optimize the weighted F1-score. Each run allows up to 300 training epochs, with a reduction factor of 3. An 80/20 train-validation split is used, and early stopping is applied to prevent overfitting. For GCN-based hypermodels, we use Optuna for more granular control over search and pruning. Each trial runs up to 300 epochs with a patience threshold of 30, and we perform 200 trials per model configuration. Search objectives again vary by dataset type, matching those used for LSTM models. The best hyperparameter set for both models is selected based on the top-performing trial, and final training proceeds for 300 full epochs to assess generalization beyond the peak epoch.
\section{Results}
\subsection{Imbalanced Dataset (Patients)}
\subsubsection{Performance Evaluation}
Table~\ref{tab:confusion} reports class-wise and aggregate metrics for the baseline (B) and duration-aware (D) variants of both LSTM and GCN models. Weighted F1-score (WF1) is the main metric, capturing class imbalance more effectively than accuracy.
\vspace{-0.8cm}
\begin{table}[htb!]
	\begin{threeparttable}
		\setlength{\tabcolsep}{1.5pt}
		\centering
		\caption{Classification Report of LSTM and GCN Models for Patients Dataset}
		\label{tab:confusion}%
		\notesize
\begin{tabularx}{\linewidth}{cccccccccccccX}
	\toprule
	Class & \multicolumn{3}{>{\hsize=\dimexpr3\hsize}c}{\textbf{B-LSTM}}
	& \multicolumn{3}{>{\hsize=\dimexpr3\hsize}c}{\textbf{D-LSTM}} 
	& \multicolumn{3}{>{\hsize=\dimexpr3\hsize}c}{\textbf{B-GCN}}
	& \multicolumn{3}{>{\hsize=\dimexpr3\hsize}c}{\textbf{D-GCN}} 
	& S \\ \cmidrule[0.8pt](lr){2-4}\cmidrule[0.8pt](lr){5-7}\cmidrule[0.8pt](lr){8-10} \cmidrule[0.8pt](lr){11-13}  \midrule 
	0   & 1      & 1      & 1      & 1      & 1      & 1      & 1      & 1      & 1      & 1      & 1      & 1      & 92     \\                                 
	1   & 0.8095 & 0.9770 & 0.8854 & 0.8047 & 0.9943 & 0.8895 & 0.7838 & 1      &0.8788  &  0.7793 & 0.9943 &0.8737 & 174    \\                     
	2   & 0.7143 & 1      & 0.8333 & 1      & 1      & 1      & 1      & 1      & 1      & 1  & 1 & 1 & 5      \\                                    
	3   & 1      & 0.9048 & 0.9500 & 1      & 1      & 1      & 1 &0.9048 &0.95 & 0.9091 & 0.9524 & 0.9302 & 21     \\                                   	
	4   & 0.7111 & 1      & 0.8312 & 0.7692 & 0.9375 & 0.8451 &  0.9143 &1 &0.9552 &0.9091 &0.9375 &0.9231 & 32     \\    
	5   & 1      & 0.5288 & 0.6918 & 1      & 0.5385 & 0.7000 &  0.9636 &0.5096 &0.6667 & 0.9815 & 0.5096 &0.6709 & 104    \\    
	Acc &        &        & \textbf{0.8715} &        &        &\textbf{0.8808} &        &        & \textbf{0.8762} &        &        & \textbf{0.8715} & 428    \\                                  
	MF1 & 0.8725 & 0.9018 & 0.8653 & 0.9290 & 0.9117 & 0.9058 &  0.9436&0.9024 &0.9084 & 0.9298 &0.899 &0.8997  & 428    \\   
	WF1 & 0.8976 & 0.8715 & \textbf{0.8615} & 0.9033 & 0.8808 & \textbf{0.8706} & 0.8969 &0.8762 & \textbf{0.8639} &  0.8945 &0.8715 &\textbf{0.8595} & 428    \\
	\bottomrule
\end{tabularx}%
		\begin{tablenotes}[flushleft]
			\scriptsize
			\item[$\bullet$]  S:Support;  MF1: Macro Average F1; WF1: Weighted Average F1;
			\item[$\bullet$]  For each model, columns are precision, recall and F1-score, respectively.
		\end{tablenotes}
	\end{threeparttable}
	\end{table}
\vspace{-0.6cm}
Overall, D-LSTM outperforms B-LSTM on both WF1 (0.8706 vs. 0.8615) and accuracy (0.8808 vs. 0.8715), indicating that incorporating pseudo-duration embeddings enhances the model’s ability to discriminate across temporally heterogeneous cases. In contrast, B-GCN marginally outperforms D-GCN on WF1 (0.8639 vs. 0.8595), with negligible difference in accuracy, suggesting that graph-based models may be less sensitive to temporal augmentation. Class-level analysis provides further insight: D-LSTM improves performance across nearly all classes. Notably, classes 2 and 3—both minority classes—see increased F1, showing that pseudo-embedding duration information stabilizes learning where raw data is sparse or noisy. For class 1 (majority), D-LSTM achieves higher recall without sacrificing precision, a critical property for reducing false negatives in outcome prediction. Class 5 remains challenging. Both duration-aware models show slightly higher F1 and little precision improvement over their baselines. Analysis of misclassifications reveals 3–5 instances consistently mislabelled as class 4, pointing to possible overlap in feature space, label noise, or distribution shift. These challenges suggest that class 5's data is either inherently ambiguous or poorly represented, limiting all models' ability to generalize effectively.

Overall, duration-aware pseudo-embeddings clearly improve the performance of sequential models such as LSTMs, but offer limited benefits in GCNs. This is likely due to architectural constraints when it comes to leveraging dynamic features over static graph structures.
\subsubsection{Hyperparameters Insights} Table~\ref{tab:Imresults} details the tuned hyperparameters for all models. B-LSTM uses a dual-layer structure with high dropout and L2 regularization to prevent overfitting from dense inputs. It employs the Adam optimizer with exponential decay, and a moderate dense output layer balancing expressiveness and generalization. D-LSTM extends this design with a dual-branch architecture: separate LSTM streams for event and duration aware pseudo-embedding inputs, merged via fusion layers. It uses RMSprop, which better handles non-stationary dynamics, and a piecewise constant learning rate schedule. A wider dense layer supports the richer fused representation, enabling better adaptation to underrepresented classes. For GCNs, tuning reveals key differences. D-GCN applies more GCN layers and wider hidden units, reflecting the need for deeper representations when fusing duration embeddings. It processes event and duration inputs via separate GCN pathways, integrating them later, implying that temporal signals require dedicated structural propagation. To manage overfitting, D-GCN uses higher dropout (up to 0.46), smaller learning rates, and lower L1 regularization. While both models use Adam, B-GCN adopts cosine decay, while D-GCN favors OneCycleLR, allowing for more dynamic training regimes. Pooling strategies also diverge: max-pooling in B-GCN vs. additive aggregation in D-GCN, possibly better preserving temporal signal accumulation. Despite this complexity, D-GCN's performance does not consistently exceed its baseline, reinforcing that static GCNs struggle to exploit temporal augmentations compared to their sequential counterparts.
\vspace{-0.2cm}
\subsection{Balanced Dataset (BPI12)}
\subsubsection{Performance Evaluation} Table~\ref{tab:result} reports the classification accuracy of baseline (B) and duration-aware (D) variants of LSTM and GCN models on the balanced BPIC12 dataset. All models, regardless of architecture or temporal augmentation, achieve perfect accuracy (100\%) across the three target classes: accept, decline, and cancel. 
\vspace{-0.8cm}
\begin{table}[htb!]
\centering
\setlength{\tabcolsep}{0.8pt}
\caption{Accuracy Scores of LSTM and GCN HyperModels and Previous Research Models on BPIC12 Dataset}
\label{tab:result}
\notesize
\begin{tabularx}{\linewidth}{Xccccccccccc}
\hline
 & SVM\cite{teinemaa2019outcome} & LR\cite{teinemaa2019outcome} & RF\cite{teinemaa2019outcome} & XGB\cite{teinemaa2019outcome} & LSTM\cite{wang2019outcome} & CNN\cite{pasquadibisceglie2020orange} & DT\cite{donadello2023outcome}   & \textbf{BGCN} & \textbf{DGCN} & \textbf{BLSTM} \textbf{DLSTM}\\
    accept & 0.63 & 0.65 & 0.69 & 0.7  & 0.71 & 0.67 & 1    & \textbf{1}    & \textbf{1}    & \textbf{1} & \textbf{1}\\
    decline & 0.55 & 0.59 & 0.6  & 0.62 & 0.64 & 0.61 & 1    & \textbf{1}    & \textbf{1}    & \textbf{1} & \textbf{1} \\
    cancel & 0.70 & 0.69 & 0.7  & 0.7  & 0.73 & 0.7  & 1    & \textbf{1}    & \textbf{1}    & \textbf{1} & \textbf{1}\\
    avg & 0.63 & 0.64 & 0.66 & 0.67 & 0.69 & 0.66 & 1    & \textbf{1}    & \textbf{1}    & \textbf{1} & \textbf{1}\\
\bottomrule
\end{tabularx}
\end{table}
\vspace{-0.5cm}
Unlike prior approaches that decompose outcome prediction into multiple binary classifiers~\cite{wang2019outcome}, our models use a single multiclass classifier per instance. This simplifies deployment while achieving comparable or better performance than more complex pipelines, such as decision tree ensembles~\cite{donadello2023outcome}. Duration-aware variants (D-LSTM, D-GCN) match baseline performance, showing that while temporal augmentation is supported, it is not necessary in settings where input traces are highly regular and easily separable.

This outcome is expected given the structural simplicity of BPIC12. Its low-variance, rigid traces enable even standard models to generalize effectively. While this supports model correctness, it provides limited evidence of robustness or temporal generalization. More challenging datasets with temporal irregularities are needed for stronger validation.
\subsubsection{Hyperparameters Insights} All four models—B-LSTM, D-LSTM, B-GCN, and D-GCN—achieve perfect accuracy on the balanced BPIC12 dataset, yet their architectural responses to temporal signals differ significantly. B-LSTM, lacking explicit duration encoding, requires deeper recurrent layers to model variation, while D-LSTM benefits from dual stream inputs with pseudo-embeddings, achieving similar results with shallower configurations. Similarly, B-GCN depends on case-level graph encoding and lean activations, whereas D-GCN leverages separate GCN streams for duration and event semantics, employing deeper post GCN layers and stronger regularization. Optimization strategies also diverge, with B-models favoring simpler schedulers and D-models adapting to overfitting risk. These patterns highlight how duration-aware inputs not only enhance learning but also influence model efficiency and robustness.
\vspace{-0.3cm}
\section{Conclusion}
This paper presents a systematic investigation of temporal representation strategies for outcome-oriented PPM, leveraging recurrent and graph-based neural architectures. We propose a dual-input model design that separately encodes event-level and case-level attributes and introduce a duration-aware pseudo-embedding mechanism to explicitly capture temporal irregularities across events.

To evaluate the effectiveness of this approach, two duration-augmented model variants are developed: D-LSTM and D-GCN, which enhance their respective baselines by incorporating time-aware representations. These models demonstrate improved handling of heterogeneous timestamp distributions by explicitly modelling temporal dynamics, particularly in event sequences with irregular or sparse timing patterns. To enable fair and robust comparisons across different input granularities and datasets, we implement self-tuned hypermodels. Empirical evaluations on balanced and imbalanced datasets demonstrate that the duration-aware variants consistently match or exceed the performance of their baseline counterparts while maintaining leaner model complexity and enhanced interpretability. Our findings reveal two key insights: (i) Augmenting event-level inputs with duration-informed features reduces reliance on deep or complex architectural designs by enabling more efficient learning of temporal dependencies. (ii) The dual-input architecture facilitates better generalisation across diverse process contexts by decoupling structural and temporal information.

By aligning the model architecture with the temporal characteristics inherent in event logs, this work provides a flexible and semantically grounded framework for predictive monitoring. Future research will extend this framework to accommodate noisy, multisource and streaming environments \cite{wang2025time}, critical steps toward scaling up learning-based techniques for real-world PPM applications.

\bibliographystyle{splncs03_unsrt}
\bibliography{ref}

\appendix
\renewcommand{\thetable}{\thesection.\arabic{table}}
\setcounter{table}{0}
\section{Appendix}
\begin{sidewaystable}
\begin{threeparttable}
\centering
\setlength{\tabcolsep}{0.8pt}
\caption{Hyperparameter Matrix for GCN and LSTM Models}
\label{tab:Imresults}
\scriptsize
\begin{tabularx}{\linewidth}{cccccccccccccccccccccX}   
\toprule
Model   & B   & G(L) & G(U) & G(A)   & SC  & G(BM)  & G(BE)  & G(D)  & P    & D(L) & D(U) & D(A)   & D(BM)  & D(BE)  & D(D)  & Opt   & LR    & WD    & Sch   & Loss  & L1   \\
\cmidrule[1pt](lr){1-2}
\cmidrule[1pt](lr){3-9}
\cmidrule[1pt](lr){11-16}
\cmidrule[1pt](lr){17-19}
BG(I)    &       32    &    2  &    88     &    GELU     &F            &           &            &  &    max    &   2(S)$^*$    &    69      &    ReLU    &           &           &    0.1601    &    Adam    &    1.248e-3    &    7.736e-4    &    Cos &    CE    &    3.424e-4    \\
  &   (88)       &           &    151    &  l\_rl    & T          &    0.1764    &    2.471e-3    &            &           &       &    133     &    GELU    &   0.9650    &    9.671e-3    &    0.3444    &    (0.8837,    &           &           &   (3.037e-3, &           &    \\\cmidrule(lr){11-16}
 &            &           &           &           &            &            &              &            &                &   1(C)$^*$    &    188     &    GELU    &    0.2754    &    8.581e-3    &    0.3062    &    0.9405)    &           &           &   50)    &           &  \\                    \midrule
DG(I)   &   32    &   2(N)$^*$    &   245   &   GELU   &          &         &          &       &   add     &  1(S)$^*$   &   228    &  l\_rl  &         &         &         &  Adam  &   1.361e-4   &   8.761e-4  &  OCL  &   CE   &   7.324e-5  \\ \cmidrule(lr){11-16}
           &  (61)       &        &   203   &  GELU   &          &   0.5817  &   9.520e-3  &       &                  &  2(C)$^*$  &   91     &  l\_rl  &         &         &   0.1195  &  (0.9374)  &          &          & (1.831e-2, &       &  \\\cmidrule(lr){4-10}
&         & 4(P)  & 38    & GELU  &       & 0.5377 & 2.934e-3 & 0.1011 &                    &        &  232    &  ELU   &   0.4174  &   9.586e-3  &   0.1100  &   0.9305) &       &          &  0.1203, &       &        \\
&         &        &   194   &  ReLU   &          &          &            &         &          &           &        &         &         &            &           &          &           &          &  54000) &       &  \\
&         &        &   196   &  ELU    &                   &            &         &          &          &        &         &        &           &            &           &          &           &          &           &        &   \\
&       &       &   159   &  sp     &          &   0.2075  &   6.602e-3  &   0.4605  &             &       &       &       &       &       &       &       &       &       &       &       &  \\\cmidrule(lr){3-9}
&       & 1(C)$^*$  & 32    & ReLU  & T     &       &       &       &       &       &            &       &       &       &       &       &       &       &       &       &  \\
\midrule
BG(12) &  32       &    2        &    151      &    tanh    & F     &     0.2745   &   1.149e-3    &    0.469   &    max       &      1(S)$^*$       &      111        &      tanh       &                 &                 &        &       RMS &       7.234e-4       &       4.415e-3       &       Cy &       MM       &       5.292e-4     \\\cmidrule(lr){11-16}
&      (25)      &       & 128   &    sp & F     &   0.4512   &  6.046e-3  &          &                & 1(C)$^*$  &       169          &       l\_rl &                 &       &  0.243   & (0.9111, &   0.7603,         & 8.523e-8)                 &  (5.0392,     & 25, & 0.0019) \\\midrule
DG(12)    &     64     &  2(N)$^*$     &  48     &  l\_rl   &            &           &         &         &       add     &    2(S)$^*$     &     48       &    ELU     &             &             &     0.2663    &     Adam     &     2.459e-3     &     2.109e-3     &     RP &     CE     &     3.930e-4    \\
&    (10)     &  &  218    &  GELU    &       &  0.8015   &  4.175e-3  &  0.2708    &            &       &     128      &    ELU     &             &             &             &     (0.9692,    &             &             &    (Max, &             &    \\ \cmidrule(lr){3-9} \cmidrule(lr){11-16}
&             & 3(P)$^*$  &  128    &  ReLU    &       &  0.2464   &  5.780e-3  &       &       &             1(C)$^*$     &     83       &    ReLU     &             &             &     0.3718    &           0.9222) &             &             &  0.8003,  &             &  \\
&             &          & 156   & ELU  &       &  0.3509   &  4.178e-4  &         &       &            &                  &             &             &             &             &         &             &             & 25,   &             &  \\
 &             &          &  228    &  l\_rl  &       &  0.2953   &  8.932e-3  &          &            &             &             &             &             &             &             &         &             &             & 7.531e-3, &             &  \\
 \cmidrule(lr){3-9} 
&       & 1(C)$^*$  & 128   & ELU  & T      &       &         & 0.2996 &       &           &       &       &       &       &       &       &       &       &      4.367e-3) &       &   \\\toprule
M      & B   & L(L)  & L(U) & \multicolumn{2}{c}{L(L2)}     & L(BM) & L(BE)     & L(D)   &  & D(L) & D(U) & D(A)    & \multicolumn{2}{c}{D(L2)}     & D(D)   & Opt    & LR        &  &  &  &  \\\cmidrule[1pt](lr){1-2}
\cmidrule[1pt](lr){3-9}
\cmidrule[1pt](lr){11-16}
\cmidrule[1pt](lr){17-18}
BL(I)  & 32  & 2     & 160  & \multicolumn{2}{c}{1.956e-4} & 0.81  & 3.345e-4 & 0.4914 &  & 1    & 144  & ReLU    & \multicolumn{2}{c}{2.017e-4} & 0.4581 & (Adam  & Exp       &  &  &  &  \\
       &     &       & 48   & \multicolumn{2}{c}{4.433e-3} &       &           & 0.3156 &  &      &      &         & \multicolumn{1}{l}{}    &     & 0.93   & 0.992) & 2.718E-03 &  &  &  &  \\\midrule
DL(I)  & 16  & 1(N)* & 256  & \multicolumn{2}{c}{1.265e-3} & 0.61  & 6.736e-4 & 0.2088 &  & 2    & 192  & (l\_rl  & \multicolumn{2}{c}{2.857e-3} & 0.4401 & rms    & P-C       &  &  &  &  \\\cmidrule(lr){3-9} 
       &     & 2(P)$^*$  & 256  & \multicolumn{2}{c}{2.990e-3} &       &           & 0.4085 &  &      & 256  & 0.1997) & \multicolumn{2}{c}{9.855e-5} & 0.2622 &        & 5.480e-4 &  &  &  &  \\
       &     &       & 64   & \multicolumn{2}{c}{9.411e-3} & 0.11  & 2.592e-5 & 0.3875 &  &      &      & ReLU    & \multicolumn{1}{l}{}    &     &        &        &           &  &  &  &  \\\cmidrule(lr){3-9}
       &     & 2(C)$^*$  & 128  & \multicolumn{2}{c}{1.121e-4} &       &           & 0.3635 &  &      &      &         & \multicolumn{1}{l}{}    &     &        &        &           &  &  &  &  \\
       &     &       & 96   & \multicolumn{2}{c}{1.140e-4} & 0.21  & 3.468e-4 & 0.4356 &  &      &      &         & \multicolumn{1}{l}{}    &     &        &        &           &  &  &  &  \\\midrule
BL(12) & 16  &       & 160  & \multicolumn{2}{c}{1.351e-5} & 0.01  & 1.022e-5 & 0.3449 &  & 3    & 80   & (l\_rl  & \multicolumn{2}{c}{1.868e-3} & 0.1946 & rms    & Exp       &  &  &  &  \\
       &     &       & 224  & \multicolumn{2}{c}{1.011e-5} &       &           & 0.2    &  &      & 16   & 0.01)   & \multicolumn{2}{c}{1.001e-5} & 0.1    &        & 7.933e-3 &  &  &  &  \\
       &     &       &      & \multicolumn{1}{l}{}    &     &       &           &        &  &      & 16   & ReLU    & \multicolumn{2}{c}{1.012e-4} & 0.1123 &        &           &  &  &  &  \\\midrule
DL(12) & 128 & 1(N)* & 224  & \multicolumn{2}{c}{1.410e-3} &       &           & 0.3647 &  & 1    & 224  & tahn    & \multicolumn{2}{c}{1.098e-4} & 0.5757 & (Adam  & Poly      &  &  &  &  \\\cmidrule(lr){3-9}
       &     & 1(P)$^*$  & 32   & \multicolumn{2}{c}{2.103e-5} & 0.81  & 2.403e-3 & 0.3106 &  &      &      &         & \multicolumn{1}{l}{}    &     &        & 0.91   &           &  &  &  &  \\\cmidrule(lr){3-9}
       &     & 1(C)$^*$  & 160  & \multicolumn{2}{c}{1.567e-5} & 0.51  & 8.522e-5 & 0.3911 &  &      &      &         & \multicolumn{1}{l}{}    &     &        & 0.991) &           &  &  &  &  \\\bottomrule
\end{tabularx}
\begin{tablenotes}
	\scriptsize
	\item[$\bullet$] Model: BG/BL/DG/DL: B-GCN/B-LSTM/D-GCN/D-LSTM; I: Patients Dataset; 12: BPI12 Dataset
	\item[$\bullet$] B: Batch size (Best Epoch); G(L)/L(L)/D(L): Number of hidden G(GCN)/L(LSTM)/D(Dense) layers; G(U)/L(U)/D(U): Units; G(A)/D(A): Activation; G(BE)/L(BE)/D(BE): Batch normalization epsilon; G(BM)/L(BM)/D(BM): Batch normalization momentum;  G(D)/L(D)/D(D): Dropout rates; SC: Skip Connection flag: P: Pooling Method; Opt:Optimizer; LR: Learning Rate; WD: Weight Decay; Sch: Learning Rate Scheduler; Loss: Loss function; L1/L2: L1/L2 regularize; Empty cell in (BM)/(BE)/(D): No batch normalization or dropout applied. 
	\item[$\bullet$] $*$:(N): Node/Event input (P):Pseudo-embedding input ; (S): Graph/case input; (C): Concatenation layer; 
	\item[$\bullet$] l\_rl: Leary\_ReLU; sp:softplus; CE: CrossEntropy; MM: MultiMargin.
	\item[$\bullet$] Adam: Adam ($\beta_1$, $\beta_2$); SGD: SGD(momentum); rms:RMSprop($\alpha$, momentum, eps); Step: Step(step size, $\gamma$); Exp: Exponential($\gamma$); RP: Reduce-on-Plateau(factor, patience, threshold, eps); Poly: Polynomial(total\_iters, power); Cos: Cos(eta\_min, T\_max); OCL: One Cycle(max, pct\_start, total\_steps); Cy: Cyclic(base, max, step\_size\_up)
    \end{tablenotes}
    \end{threeparttable}
\end{sidewaystable}

\end{document}